\pdfoutput=1

\documentclass[11pt]{article}

\usepackage{ACL2023}

\usepackage{times}
\usepackage{latexsym}

\usepackage{header}

\usepackage[T1]{fontenc}

\usepackage[utf8]{inputenc}

\usepackage{microtype}

\usepackage{inconsolata}
\usepackage{arydshln}
\usepackage{booktabs}
\usepackage{adjustbox}
\usepackage{soul}
\usepackage{amsmath}

\newcommand{\dataset}{\textsc{Quest}}

\makeatletter
\renewcommand{\paragraph}{%
  \@startsection{paragraph}{4}{\z@}{0.5ex plus%
   0.5ex minus .2ex}{-1em}{\normalsize\bf}%
}
\makeatother

\title{\dataset: A Retrieval Dataset of Entity-Seeking Queries\\ with Implicit Set Operations}

\author{
    Chaitanya Malaviya\textsuperscript{1}\thanks{\,\,\,Work done during an internship at Google.},
    Peter Shaw\textsuperscript{2},
    Ming-Wei Chang\textsuperscript{2},
    Kenton Lee\textsuperscript{2},
    Kristina Toutanova\textsuperscript{2}\\
    \\
    \textsuperscript{1}University of Pennsylvania
    \textsuperscript{2}Google DeepMind\\
    {\tt cmalaviy@seas.upenn.edu}\\
    {\tt \{petershaw,mingweichang,kentonl,kristout\}@google.com}
}

\begin{document}
\maketitle

\begin{abstract}
Formulating selective information needs results in queries that implicitly specify set operations, such as intersection, union, and difference. For instance, one might search for "shorebirds that are not sandpipers" or "science-fiction films shot in England". To study the ability of retrieval systems to meet such information needs, we construct \dataset, a dataset of 3357 natural language queries with implicit set operations, that map to a set of entities corresponding to Wikipedia documents. The dataset challenges models to match multiple constraints mentioned in queries with corresponding evidence in documents and correctly perform various set operations. The dataset is constructed semi-automatically using Wikipedia category names. Queries are automatically composed from individual categories, then paraphrased and further validated for naturalness and fluency by crowdworkers. Crowdworkers also assess the relevance of entities based on their documents and highlight attribution of query constraints to spans of document text.  We analyze several modern retrieval systems, finding that they often struggle on such queries. Queries involving negation and conjunction are particularly challenging and systems are further challenged with combinations of these operations.\footnote{The dataset is available at \url{https://github.com/google-research/language/tree/master/language/quest}.}
\end{abstract}

\section{Introduction}

\begin{figure}[t!]
    \centering
    \includegraphics[width=\columnwidth,height=7cm,keepaspectratio]{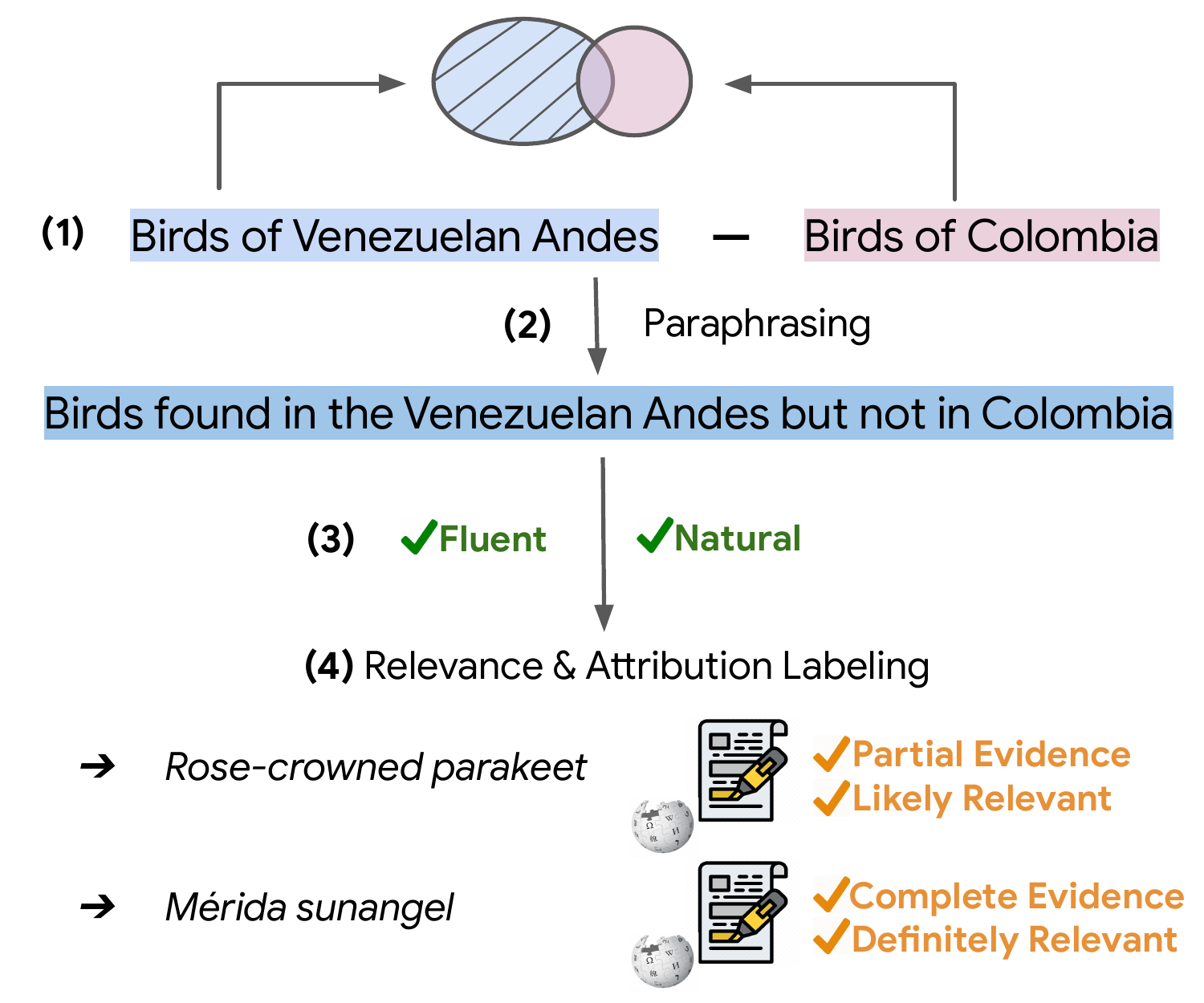}
    \caption{The dataset construction process for \dataset. First, (1) we sample Wikipedia category names and find their corresponding set of relevant entities. (2) Then, we compose a query with set operations and have this query paraphrased by crowdworkers. (3) These queries are then validated for fluency and naturalness. (4) Finally, crowdworkers mark the entities' relevance by highlighting attributable spans in their documents.}
    \label{fig:quest_construction}
\end{figure}

People often express their information needs with multiple preferences or constraints. Queries corresponding to such needs typically implicitly express set operations such as intersection, difference, and union. For example, a movie-goer might be looking for a \textit{science-fiction film from the 90s which does not feature aliens} and a reader might be interested in a \textit{historical fiction novel set in France}. Similarly, a botanist attempting to identify a species based on their recollection might search for \textit{shrubs that are evergreen and found in Panama}. Further, if the set of entities that satisfy the constraints is relatively small, a reader may like to see and explore an exhaustive list of these entities. In addition, to verify and trust a system's recommendations, users benefit from being shown evidence from trusted sources \cite{10.1162/tacl_a_00398}.

Addressing such queries has been primarily studied in the context of question answering with structured knowledge bases (KBs), where query constraints are grounded to predefined predicates and symbolically executed. However, KBs can be incomplete and expensive to curate and maintain.
Meanwhile, advances in information retrieval  may enable developing systems that can address such queries without relying on structured KBs, by matching query constraints directly to supporting evidence in text documents. However, queries that combine multiple constraints with implicit set operations are not well represented in existing retrieval benchmarks such as MSMarco \cite{nguyen2016ms} and Natural Questions \cite{kwiatkowski-etal-2019-natural}. Also, such datasets do not focus on retrieving an exhaustive document set, instead limiting annotation to the top few results of a baseline information retrieval system.

To analyze retrieval system performance on such queries, we present \dataset, a dataset with natural language queries from four domains, that are mapped to relatively comprehensive sets of entities corresponding to Wikipedia pages.  We use categories and their mapping to entities in Wikipedia as a building block for our dataset construction approach, but do not allow access to this semi-structured data source at inference time, to simulate text-based retrieval. Wikipedia categories represent a broad set of natural language descriptions of entity properties and often correspond to selective information need queries that could be plausibly issued by a search engine user. The relationship between property names and document text is often subtle and requires sophisticated reasoning to determine, representing the natural language inference challenge inherent in the task.

Our dataset construction process is outlined in Figure~\ref{fig:quest_construction}. The base queries are semi-automatically generated using Wikipedia category names. To construct complex queries, we sample category names and compose them  by using pre-defined templates (for example, $A \cap B \setminus C$). 
Next, we ask crowdworkers to paraphrase these automatically generated queries, while ensuring that the paraphrased queries are fluent and clearly describe what a user could be looking for. These are then validated for naturalness and fluency by a different set of crowdworkers, and filtered according to those criteria. Finally, for a large subset of the data, we collect scalar relevance labels based on the entity documents and fine-grained textual attributions mapping query constraints to spans of document text. Such annotation could aid the development of systems that can make precise inferences from trusted sources. 

\begin{figure}[t!]
    \centering
    \includegraphics[width=\columnwidth]{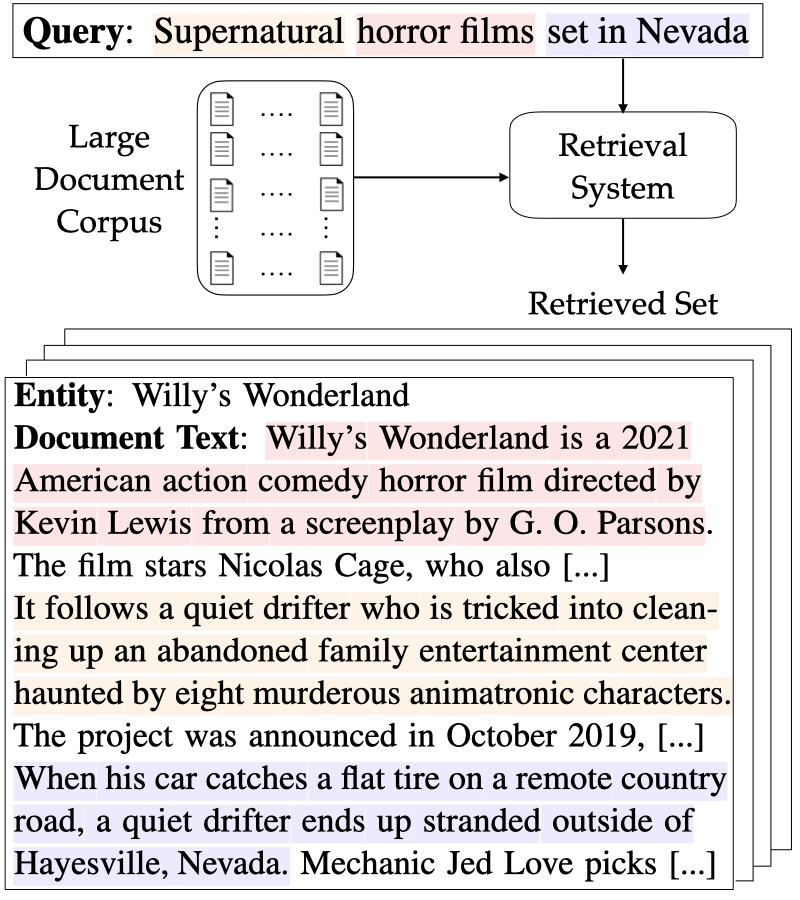}
    \caption{An example of a query and relevant entity from \dataset. The attribution for different query constraints can come from different parts of the document.}
    \label{fig:quest-example}
\end{figure}

Performing well on this dataset requires systems that can match query constraints with corresponding evidence in documents and handle set operations implicitly specified by the query (see Figure~\ref{fig:quest-example}), while also efficiently scaling to large collections of entities. We evaluate several retrieval systems by finetuning pretrained models on our dataset. Systems are trained to retrieve multi-document sets given a query. We find that current dual encoder and cross-attention models up to the size of T5-Large~\cite{raffel2019exploring} are largely not effective at performing retrieval for queries with set operations. Queries with conjunctions and negations prove to be especially challenging for models and systems are further challenged with combinations of set operations. Our error analysis reveals that non-relevant false positive entities are often caused by the model ignoring negated constraints, or ignoring the conjunctive constraints in a query.

\section{Related Work}

Previous work in question answering and information retrieval has focused on QA over knowledge bases as well as open-domain QA and retrieval over a set of entities or documents. We highlight how these relate to our work below.

\paragraph{Knowledge Base QA}

Several datasets have been proposed for question answering over knowledge bases \cite[\textit{inter alia}]{berant-etal-2013-semantic,yih-etal-2016-value,talmor-berant-2018-web,keysers2020measuring,gu2021beyond}. These benchmarks require retrieval of a set of entities that exist as nodes or relations in an accompanying knowledge base. Questions are optionally supplemented with logical forms.
\citet{lan2021survey} provide a comprehensive survey of complex KBQA datasets. 

Previous work has simultaneously noted that large curated KBs are incomplete \cite{watanabe2017question}. Notably, KBQA systems operate over a constrained answer schema, which limits the types of queries they can handle. Further, these schema are expensive to construct and maintain. For this reason, our work focuses on a setting where we do not assume access to a KB. We note that KBQA datasets have also been adapted to settings where a KB is incomplete or unavailable \cite{watanabe2017question, sun-etal-2019-pullnet}. This was done by either removing some subset of the data from the KB or ignoring the KB entirely. A key difference from these datasets is also that we do not focus on multi-hop reasoning over multiple documents. Instead, the relevance of an entity can be determined solely based on its document.

\paragraph{Open-Domain QA and Retrieval} Many open-domain QA benchmarks, which consider QA over unstructured text corpora, have been proposed in prior work. Some of these, such as TREC \cite{craswell2020overview}, MSMarco \cite{nguyen2016ms} and Natural Questions \cite{kwiatkowski-etal-2019-natural} are constructed using "found data", using real user queries on search engines. \citet{thakur2021beir} present a benchmark where they consider many such existing datasets. Datasets such as HotpotQA \cite{yang-etal-2018-hotpotqa}, and MultiRC \cite{khashabi-etal-2018-looking} have focused on multi-hop question answering. Other work has explored e-commerce datasets (for example, \cite{kong2022multi}), but these have not been released publicly. Notably, the focus of these datasets differs from ours as we focus on queries that contain implicit set operations over exhaustive answer sets. Such queries are not well represented in existing datasets because they occur in the tail of the query distributions considered.

\paragraph{Multi-Answer Retrieval}

Related work \cite{min-etal-2021-joint, amouyal2022qampari} also studies the problem of \emph{multi-answer retrieval}, where systems are required to predict multiple distinct answers for a query. \citet{min-etal-2021-joint} adapt existing datasets (for example, WebQuestionsSP \cite{yih-etal-2016-value}) to study this setting and propose a new metric, MRecall@K, to evaluate exhaustive recall of multiple answers. We also consider the problem of  multi-answer set retrieval, but consider queries that implicitly contain set constraints.

In concurrent work, RomQA~\cite{zhong2022romqa} proposes an open-domain QA dataset, focusing on combinations of constraints extracted from Wikidata. RomQA shares our motivation to enable answering queries with multiple constraints, which have possibly large answer sets. To make attribution to evidence feasible without human annotation, RomQA focuses on questions whose component constraints can be verified from single entity-linked sentences from Wikipedia abstracts, annotated with relations automatically through distant supervision, with high precision but possibly low recall (T-Rex corpus). In \textsc{Quest}, we broaden the scope of query-evidence matching operations by allowing for attribution through more global, document-level inference. To make human annotation for attribution feasible, we limit the answer set size and the evidence for an answer to a single document.

\section{Dataset Generation}
\label{sec:dataset}

\begin{table*}[ht!]
\centering
\footnotesize
\scalebox{.85}{
\begin{tabular}{ |c|c|c|c| } \hline
\textbf{Domain} & \textbf{Template} & \textbf{Example} & \textbf{Num. Queries} \\ \hline
 & $A$ & Biographical Italian bandits films & 125 \\
 & $A \cup B$ & Dutch crime comedy or romantic comedy films & 135 \\
 & $A \cap B$ & Italian crime films set in the 1970's & 143 \\
 Films & $A \setminus B$ & Indian sport films that are not about cricket & 126 \\
 & $A \cup B \cup C$ & Dutch or Swiss war films, or war films from 1945 & 122 \\
 & $A \cap B \cap C$ & 2020's drama films shot in cleveland & 124 \\
 & $A \cap B \setminus C$ & Epic films about Christianity not set in Israel & 121 \\
\hdashline
 & $A$ & 2004 German novels & 125 \\
 & $A \cup B$ & 1925 Russian novels or Novels by Ivan Bunin & 125 \\
 & $A \cap B$ & 1991 Novels set in Iceland & 133 \\
 Books & $A \setminus B$ & Novels set in the 1900s not based on real events & 123 \\
 & $A \cup B \cup C$ & Novels set in Nanjing, Hebei, or Jiangsu & 125 \\
 & $A \cap B \cap C$ & English language Harper \& Brothers Children's fiction books & 124 \\
 & $A \cap B \setminus C$ & Novels that take place in Vietnam that aren't about war & 115 \\
\hdashline
 & $A$ & plants only from Gabon & 115 \\
 & $A \cup B$ & Trees of Manitoba or Subarctic America & 125 \\
 & $A \cap B$ & Shrubs used in traditional Native American medicine & 135 \\
 Plants & $A \setminus B$ & Trees from the Northwestern US that can't be found in Canada & 61 \\
 & $A \cup B \cup C$ & Moths or Insects or Arthropods of Guadeloupe & 121 \\
 & $A \cap B \cap C$ & Plants the Arctic, the United Kingdom, and the Caucasus have in common & 123 \\
 & $A \cap B \setminus C$ & Orchids of Indonesia and Malaysia but not Thailand & 122 \\
 \hdashline
 & $A$ & what are the Rodents of Cambodia & 115 \\
 & $A \cup B$ & Animals from Cuba or Jamaica that are extinct & 121 \\
 & $A \cap B$ & Neogene mammals of Africa that are Odd-toed ungulates & 111 \\
 Animals & $A \setminus B$ & Non-Palearctic birds of Mongolia & 110 \\
 & $A \cup B \cup C$ & Cenozoic birds of Asia or Africa or Paleogene birds of Asia & 114 \\
 & $A \cap B \cap C$ & Birds of Chile that are also Birds of Peru and Fauna of the Guianas & 104 \\
 & $A \cap B \setminus C$ & mammals found in the Atlantic Ocean and Colombia, but not in Brazil & 114 \\ \hline
\end{tabular}
}
\caption{Templates used for construction of queries with set operations and examples from the four domains considered, along with the count of examples per each domain and template.} 
\label{tab:examples}
\end{table*}

\textsc{Quest} consists of 3357 queries paired with up to 20 corresponding entities. Each entity has an associated document derived from its Wikipedia page. The dataset is divided into 1307 queries for training, 323 for validation, and 1727 for testing.

The task for a system is to return the correct set of entities for a given query. Additionally, as the collection contains 325,505 entities, the task requires retrieval systems that can scale efficiently. We do not allow systems to access additional information outside of the text descriptions of entities at inference time. Category labels are omitted from all entity documents.

\subsection{Atomic Queries}
The base atomic queries (i.e., queries without any introduced set operations) in our dataset are derived from Wikipedia category names\footnote{We
use the Wikipedia version from 06/01/2022.}. These are hand-curated natural language labels assigned to groups of related documents in Wikipedia\footnote{Note that these category labels can sometimes be conjunctive themselves, potentially increasing complexity.}. Category assignments to documents allow us to automatically determine the set of answer entities for queries with high precision and relatively high recall. We compute transitive closures of all relevant categories to determine their answer sets.

However, repurposing these categories for constructing queries poses challenges: 1) lack of evidence in documents: documents may not contain sufficient evidence for judging their relevance to a category, potentially providing noisy signal for relevance attributable to the document text, 2) low recall: entities may be missing from categories to which they belong. For about half of the dataset, we crowdsource relevance labels and attribution based on document text, and investigate recall through manual error analysis (\S\ref{sec:exp-main-results}).

We select four domains to represent some diversity in queries: films, books, animals and plants. Focusing on four rather than all possible domains enables higher quality control. The former two model a general search scenario, while the latter two model a scientific search scenario.

\begin{table*}[t!]
\centering
\scalebox{0.82}{
\begin{tabular}{lccccc}
\toprule
& Films & Books & Plants & Animals & \bf All \\

\midrule
Num. Queries & 896 & 870 & 802 & 789 & 3357 \\
Num. Entities & 146368 & 50784 & 83672 & 44681 & 325505 \\
Avg. Query Len. & 8.68 & 7.93 & 8.94 & 9.09 & 8.64 \\
Avg. Doc. Len. & 532.2 & 655.3 & 258.1 & 293.1 & 452.2 \\
Avg. Ans. Set Size & 8.8 & 8.6 & 12.2 & 12.6 & 10.5 \\
\bottomrule
\end{tabular}}
\caption{Statistics of examples in \textsc{Quest} across different domains.}
\label{tab:dataset_stats}
\end{table*}

\subsection{Introducing set operations}

To construct queries with set operations, we define templates that represent plausible combinations of atomic queries. Denoting atomic queries as A, B and C, our templates and corresponding examples from different domains are listed in Table~\ref{tab:examples}. Templates were constructed by composing three basic set operations (intersection, union and difference). They were chosen to ensure unambiguous interpretations of resulting queries by omitting those combinations of set operations that are non-associative. 

Below we describe the logic behind sampling atomic queries (i.e., $A$, $B$, $C$) for composing complex queries, with different set operations. In all cases, we ensure that answer sets contain between 2-20 entities so that crowdsourcing relevance judgements is feasible. We sample 200 queries per template and domain, for a total of 4200 initial queries. The dataset is split into train + validation (80-20 split) and testing equally. In each of these sets, we sampled an equal number of queries per template.

\paragraph{Intersection.} The intersection operation for a template $A \cap B$ is particularly interesting and potentially challenging when both $A$ and $B$ have large answer sets but their intersection is small. We require the minimum answer set sizes of each $A$ and $B$ to be fairly large (>50 entities), while their intersection to be small (2-20 entities).
\paragraph{Difference.}
Similar to intersection, we require the answer sets for both $A$ and $B$ to be substantial (>50 entities), but also place maximum size constraints on both $A$ (<200 entities) and $B$ (<10000 entities) as very large categories tend to suffer from recall issues in Wikipedia.
We also limit the intersection of $A$ and $B$ (see reasoning in Appendix~\ref{app:sets_analysis}).
\paragraph{Union.} For the union operation, we require both $A$ and $B$ to be well-represented through the entities in the answer set for their union $A \cup B$. Hence, we require both $A$ and $B$ to have at least 3 entities. Further, we require their intersection to be non-zero but less than 1/3rd of their union. This is so that $A$ and $B$ are somewhat related queries.

\noindent For all other templates that contain compositions of the above set operations, we apply the same constraints recursively. For example, for $A \cap B \setminus C$, we sample atomic queries $A$ and $B$ for the intersection operation, then sample $C$ based on the relationship between $A \cap B$ and $C$.

\subsection{Annotation Tasks}

Automatically generating queries based on templates results in queries that are not always fluent and coherent. Further, entities mapped to a query may not actually be relevant and don’t always have attributable evidence for judging their relevance. We conduct crowdsourcing to tackle these issues. The annotation tasks aim at ensuring that 1) queries are fluent, unambiguous and contain diverse natural language logical connectives, (2) entities are verified as being relevant or non-relevant and (3) relevance judgements are attributed to document text for each relevant entity. Crowdsourcing is performed in three stages, described below. More annotation details and the annotation interfaces can be found in Appendix~\ref{app:annotation}.

\subsubsection{Paraphrasing}
 Crowdworkers were asked to paraphrase a templatically generated query so that the paraphrased query is fluent, expresses all constraints in the original query, and clearly describes what a user could be looking for. This annotation was done by one worker per query.

\subsubsection{Validation}
This stage is aimed at validating the queries we obtain from the paraphrasing stage. Crowdworkers were given queries from the first stage and asked to label whether the query is 1) fluent, 2) equivalent to the original templatic query in meaning, and 3) rate its naturalness (how likely it is to be issued by a real user). This annotation was done by 3 workers per query. We excluded those queries which were rated as not fluent, unnatural or having a different meaning than the original query, based on a majority vote. Based on the validation, we removed around around 11\% of the queries from stage 1.

\begin{figure*}[t!]
    \centering
    \includegraphics[height=3cm,keepaspectratio]{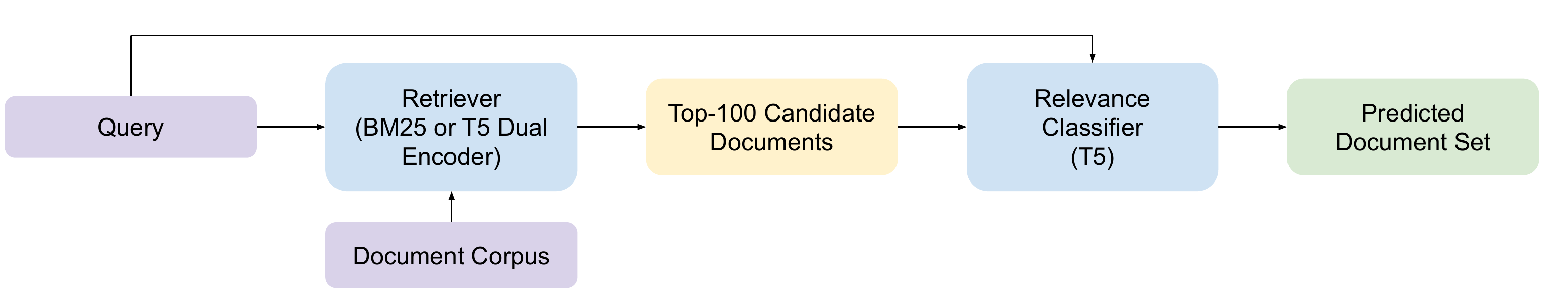}
    \caption{We compare several systems consisting of a retriever for efficiently selecting a set of candidates from the document corpus and a document relevance classifier for determining the final predicted document set.}
    \label{fig:baseline-systems}
\end{figure*}

\subsubsection{Relevance Labeling}

Next, crowdworkers were asked to provide relevance judgements for the automatically determined answer sets of queries.
Specifically, they were given a query and associated entities/documents, and asked to label their relevance on a scale of 0-3 (definitely not relevant, likely not relevant, likely relevant, definitely relevant). They were asked to ensure that relevance should mostly be inferred from the document, but they could use some background knowledge and do minimal research. 

We also asked them to provide attributions for document relevance. Specifically, we ask them to first label whether the document provides sufficient evidence for the relevance of the entity (complete/partial/no). Then, for different phrases in the query (determined by the annotator), we ask them to mark sentence(s) in the document that indicate its relevance. The attribution annotation is broadly inspired by \citet{ais_paper}. For negated constraints, we ask annotators to mark attributable sentences if they provide counter-evidence. Since this annotation was time-intensive, we collected these annotations for two domains (films and books). We found that relevance labeling was especially difficult for the plants and animals domains, as they required more specialized scientific knowledge. In our pilot study prior to larger scale data collection, we collected 3 relevance ratings from different annotators for 905 query and document pairs from the films domain. In 61.4\% of cases, all 3 raters judged the document to be “Definitely relevant” or “Likely relevant” or all 3 raters judged the document to be “Definitely not relevant” or “Likely not relevant”. The Fleiss’ kappa metric on this data was found to be K=0.43. We excluded all entities which were marked as likely or definitely not relevant to a query based on the document text from its answer set. Around 23.7\% of query-document pairs from stage 2 were excluded.

\subsection{Dataset Statistics}

Basic dataset statistics are reported in Table~\ref{tab:dataset_stats}.
The dataset contains more entities from the films domain, because this domain is more populated in Wikipedia. The average length of queries is 8.6 words and the average document length is 452 words.
Documents from the films and books domains are longer on average, as they often contain plots and storylines.
Around $\sim$69\% of entities have complete evidence and $\sim$30\% have partial evidence. Evidence was labeled as partial when not all phrases in the query had explicit evidence in the document (i.e., they may require background knowledge or reasoning). 
There are on average 33.2 words attributed for each entity with the maximum attribution text span ranging up to length 1837 words.
Finally, the average answer set size is 10.5 entities.

\subsection{Additional Training Examples}
\label{sec:augmentation}

Beyond the annotated data, we generated additional synthetic examples for training. We found including such examples improved model performance, and we include these examples for the experiments in \S\ref{sec:experiments}. To generate these examples, we sample 5000 atomic queries from all domains, ensuring that they do not already appear as sub-queries in any of the queries in \textsc{Quest} and use their corresponding entities in Wikipedia as their relevant entity set.

\section{Experimental Setup}
\label{sec:experiments}

We evaluate modern retrieval systems to establish baseline performances.
We also perform extensive error analysis to understand patterns of model errors and the quality of the labels in \textsc{Quest}.

\subsection{Task Definition}

We consider a corpus, $\mathcal{E}$, that contains entities across all domains in the dataset. Each entity is accompanied with a document based on its Wikipedia page. An example in our dataset consists of a query, $x$, and an annotated set of relevant entities, $y \subset \mathcal{E}$. As described in \S\ref{sec:dataset}, for all examples $|y| < 20$. Our task is to develop a system that, given $\mathcal{E}$ and a query $x$, predicts a set of relevant entities, $\hat{y} \subset \mathcal{E}$. 

\subsection{Evaluation}

Our primary evaluation metric is average $F_1$, which averages per-example $F_1$ scores. We compute $F_1$ for each example by comparing the predicted set of entities, $\hat{y}$, with the annotated set, $y$.

\begin{table*}[!ht]
\centering
\scalebox{0.82}{
\begin{tabular}{lcccccccc}
\toprule

Retriever (K=100) & Classifier & Avg. Precision & Avg. Recall & Avg. F1 \\

\midrule
BM25 & T5-Base & 0.168 & 0.160 & 0.141 \\
BM25 & T5-Large & 0.178 & 0.168 & 0.150 \\
T5-Large DE & T5-Base & 0.153 & 0.354 & 0.176 \\
T5-Large DE & T5-Large & 0.165 & 0.368 & \bf 0.192 \\

\bottomrule
\end{tabular}}
\caption{Average Precision, Recall, and F1 of baseline systems evaluated on the test set.}
\label{tab:ft_results}
\end{table*}
\begin{table*}[!ht]
\centering
\scalebox{0.82}{
\begin{tabular}{lcccccccc}
\toprule
 & \multicolumn{4}{c}{Avg. Recall@K} 
 & \multicolumn{4}{c}{MRecall@K} \\
 \cmidrule(lr){2-5} \cmidrule(lr){6-9} 
Retriever                       & 20 & 50 & 100 & 1000 & 20 & 50 & 100 & 1000 \\

\midrule
BM25 & 0.104 & 0.153 & 0.197 & 0.395 & 0.020 & 0.030 & 0.037 & 0.087 \\
T5-Base DE & 0.255 & 0.372 & 0.455 & 0.726 & 0.045 & 0.088 & 0.127 & 0.360 \\
T5-Large DE & \bf 0.265 & \bf 0.386 & \bf 0.476 & \bf 0.757 & \bf 0.047 & \bf 0.100 & \bf 0.142 & \bf 0.408 \\
\bottomrule
\end{tabular}}
\caption{Average Recall and MRecall of various retrievers.}
\label{tab:retriever_results}
\end{table*}

\subsection{Baseline Systems}

We evaluated several combinations of retrievers and classifiers, as shown in Figure~\ref{fig:baseline-systems}. 
For the retriever component, we consider a sparse BM25 retriever~\cite{robertson2009probabilistic} and a dense dual encoder retriever (denoted DE). Following \citet{ni2021large}, we initialize our dual encoder from a T5~\cite{raffel2019exploring} encoder and train with an in-batch sampled softmax loss~\cite{henderson2017efficient}.
Once we have a candidate set, we need to determine a set of relevant entities. To classify relevance of each candidate document for the given query, we consider a cross-attention model which consists of a T5 encoder and decoder.\footnote{Scores from BM25 and dual encoders trained with a softmax loss are not normalized to provide relevance probabilities for documents. We found that naively applying a global threshold to these scores to produce answer sets did not perform as well as using a classifier trained with a binary cross-entropy loss to predict document relevance.}
We train the cross-attention classifier using a binary cross-entropy loss with negative examples based on non-relevant documents in top 1,000 documents retrieved by BM25 and random non-relevant documents (similarly to \citet{nogueira2019passage}). 
As cross-attention classification for a large number of candidates is computationally expensive, we restrict BM25 and the dual encoder to retrieve 100 candidates which are then considered by the cross-attention classifier.
As our T5-based dual encoder can only efficiently accommodate up to 512 tokens, we truncate document text. We discuss the impact of this and alternatives in \S\ref{sec:exp-main-results}. 
Further, since T5 was pre-trained on Wikipedia, we investigate the impact of memorization in Appendix~\ref{app:memorization}.
Additional details and hyperparameter settings are in Appendix~\ref{app:hyperparams}.

\subsection{Manual Error Annotation}
\label{sec:error-annotation}
For the best overall system, we sampled errors and manually annotated 1145 query-document pairs from the validation set. For the retriever, we sampled relevant documents not included in the top-100 candidate set and non-relevant documents ranked higher than relevant ones. For the classifier, we sampled false positive and false negative errors made in the top-100 candidate set. This annotation process included judgements of document relevance (to assess agreement with the annotations in the dataset) and whether the document (and the truncated version considered by the dual encoder or classifier) contained sufficient evidence to reasonably determine relevance. We also annotated relevance for each constraint within a query. We discuss these results in \S\ref{sec:exp-main-results}.

\section{Results and Analysis}
\label{sec:exp-main-results}

We report the performance of our baseline systems on the test set in Table~\ref{tab:ft_results}. In this section, we summarize the key findings from our analysis of these results and the error annotation described in \S\ref{sec:error-annotation}.

\paragraph{Dual encoders outperform BM25.} As shown in Table~\ref{tab:ft_results}, the best overall system uses a T5-Large Dual Encoder instead of BM25 for retrieval. The performance difference is even more significant when comparing recall of Dual Encoders and BM25 directly. We report average recall (average per-example recall of the full set of relevant documents) and MRecall~\cite{min-etal-2021-joint} (the percentage of examples where the candidate set contains all relevant documents), over various candidate set sizes in Table~\ref{tab:retriever_results}.

\paragraph{Retrieval and classification are both challenging.} As we consider only the top-100 candidates from the retriever, the retriever's recall@100 sets an upper bound on the recall of the overall system. Recall@100 is only 0.476 for the T5-Large Dual Encoder, and the overall recall is further reduced by the T5-Large classifier to 0.368, despite achieving only 0.165 precision. This suggests that there is room for improvement from both stages to improve overall scores. As performance improves for larger T5 sizes for both retrieval and classification, further model scaling could be beneficial.

\paragraph{Models struggle with intersection and difference.}
We also analyzed results across different templates and domains, as shown in Table \ref{tab:breakdown_results}.
Different constraints lead to varying distributions over answer set sizes and the atomic categories used. Therefore, it can be difficult to interpret differences in F1 scores across templates. Nevertheless, we found the queries with set union have the highest average F1 scores. 
Queries with set intersection have the lowest average F1 scores, and queries with set difference also appear to be challenging. 

To analyze why queries with conjunction and negation are challenging, we labeled the relevance of individual query constraints (\S\ref{sec:error-annotation}), where a system incorrectly judges relevance of a non-relevant document. The results are summarized in Table~\ref{tab:error_categories}.
For a majority of false positive errors involving intersection, at least one constraint is satisfied. This could be interpreted as models incorrectly treating intersection as union when determining relevance. Similarly, for a majority of examples with set difference, the negated constraint is not satisfied. This suggests that the systems are not sufficiently sensitive to negations.

\begin{table}[!ht]
\centering
\scalebox{0.8}{
\begin{tabular}{lccccc}
\toprule
Template & Films & Books & Plants & Animals & \bf All \\

\midrule

$A$ & 0.231 & 0.436 & 0.209 & 0.214 & 0.274 \\
$A \cup B$ & 0.264 & 0.366 & 0.229 & 0.271 & 0.282 \\
 $A \cap B$ & 0.115 & 0.138 & 0.049 & 0.063 & 0.092 \\
$A \setminus B$ & 0.177 & 0.188 & 0.216 & 0.204 & 0.193 \\
$A \cup B \cup C$ & 0.200 & 0.348 & 0.306 & 0.294 & 0.287 \\
$A \cap B \cap C$ & 0.086 & 0.121 & 0.07 & 0.065 & 0.086 \\
$A \cap B \setminus C$ & 0.119 & 0.112 & 0.121 & 0.136 & 0.122 \\
\midrule

\bf All & 0.171 & 0.248 & 0.165 & 0.182 & 0.192 \\
\bottomrule
\end{tabular}}
\caption{F1 of our strongest baseline (T5-Large DE + T5-Large Classifier) across templates and domains.}
\label{tab:breakdown_results}
\end{table}

\paragraph{There is significant headroom to improve both precision and recall.}
As part of our manual error analysis (\S\ref{sec:error-annotation}), we made our own judgements of relevance and measured agreement with the relevance annotations in \dataset. As this analysis focused on cases where our best system disagreed with the relevance labels in the dataset, we would expect agreement on these cases to be significantly lower than on randomly selected query-document pairs in the dataset. Therefore, it provides a focused way to judge the headroom and annotation quality of the dataset.

For false negative errors, we judged 91.1\% of the entities to be relevant for the films and books domains, and 81.4\% for plants and animals. Notably, we collected relevance labels for the films and books domains and removed some entities based on these labels, as described in \S\ref{sec:dataset}, which likely explains the higher agreement for false negatives from these domains.
This indicates significant headroom for improving recall as defined by \dataset, especially for the domains where we collected relevance labels.

For false positive errors, we judged 28.8\% of the entities to be relevant, showing a larger disagreement with the relevance labels in the dataset. This is primarily due to entities not included in the entity sets derived from the Wikipedia category taxonomy (97.7\%), rather than entities removed due to relevance labeling. This is a difficult issue to fully resolve, as it is not feasible to exhaustively label relevance for all entities to correct for recall issues in the Wikipedia category taxonomy. Future work can use pooling to continually grow the set of relevant documents~\cite{pooling}. Despite this, our analysis suggests there is significant headroom for improving precision, as we judged a large majority of the false positive predictions to be non-relevant.
\begin{table}[t!]
\centering
\scalebox{0.82}{
\begin{tabular}{lccc|c}
\toprule
& \multicolumn{3}{c}{\# Constraints} & \\
 \cmidrule(lr){2-4} 
 & 1 & 2 & 3 & Neg. \\
\midrule
\bf{Retriever} & & & & \\
$A \cap B$  & 63.5 & 36.5 & --- & --- \\
$A \cap B \cap C$ & 56.5 & 37.0 & 6.5 & --- \\
$A \setminus B$ & 80.3 & 19.7 & --- & 59.1 \\
$A \cap B \setminus C$ & 47.6 & 40.5 & 11.9 & 26.2 \\
\midrule
\bf{Classifier} & & & & \\
$A \cap B$  & 83.3 & 16.7 & --- & --- \\
$A \cap B \cap C$ & 73.2 & 22.0 & 4.9 & --- \\
$A \setminus B$ & 81.0 & 19.1 & --- & 38.1 \\
$A \cap B \setminus C$ & 95.5 & 4.6 & 0.0 & 68.2 \\

\bottomrule
\end{tabular}
}

\caption{Analysis of false positive errors from the T5-Large classifier and cases where a non-relevant document was ranked ahead of a relevant one for the T5-Large dual encoder. For queries with conjunction, we determined the percentage of cases where 1, 2, or 3 constraints in the template were not satisfied by the predicted document (\# Constraints). For queries with negation, we measured the percentage of cases where the negated constraint (Neg.) was not satisfied.}

\label{tab:error_categories}
\end{table}

\paragraph{Truncating document text usually provides sufficient context.}
In our experiments, we truncate document text to 512 tokens for the dual encoder, and 384 tokens for the classifier to allow for the document and query to be concatenated. Based on our error analysis (\S\ref{sec:error-annotation}), out of the documents with sufficient evidence to judge relevance, evidence occurred in this truncated context 93.2\% of the time for the dual encoder, and 96.1\% of the time for the classifier. This may explain the relative success of this simple baseline for handling long documents. We also evaluated alternative strategies but these performed worse in preliminary experiments\footnote{For the dual encoder, we split documents into overlapping chunks of 512 tokens, and aggregated scores at inference \cite{dai2019deeper}. For the cross-attention model, we evaluated using BM25 to select the top-3 passages of length 128.}.
Future work can evaluate efficient transformer variants \cite{guo-etal-2022-longt5, Beltagy2020Longformer}.

\section{Conclusion}

We present \dataset, a new benchmark of queries which contain implicit set operations with corresponding sets of relevant entity documents. Our experiments indicate that such queries present a challenge for modern retrieval systems. 
Future work could consider approaches that have better inductive biases for handling set operations in natural language expressions (for example, \citet{vilnis-etal-2018-probabilistic}).
The attributions in \textsc{Quest} can be leveraged for building systems that can provide fine-grained attributions at inference time.
The potential of pretrained generative LMs and multi-evidence aggregation methods to answer set-seeking selective queries, while providing attribution to sources, can also be investigated. 

\section{Limitations}

\paragraph{Naturalness.} Since our dataset relies on the Wikipedia category names and  semi-automatically generated compositions, it does not represent an unbiased sample from a natural distribution of real search queries that contain implicit set operations. Further, we limit attention to non-ambiguous queries and do not address the additional challenges that arise due to ambiguity in real search scenarios. However, the queries in our dataset were judged to plausibly correspond to real user search needs and system improvements measured on {\dataset}  should correlate with improvements on at least a fraction of natural search engine queries with set operations.

\paragraph{Recall.} We also note that because Wikipedia categories have imperfect recall of all relevant entities (that contain sufficient evidence in their documents), systems may be incorrectly penalised for predicted relevant entities assessed as false positive. We quantify this in section~\ref{sec:exp-main-results}. We have also limited the trusted source for an entity to its Wikipedia document but entities with insufficient textual evidence in their documents may still be relevant. Ideally, multiple trusted sources could be taken into account and evidence could be aggregated to make relevance decisions.  RomQA~\cite{zhong2022romqa} takes a step in this latter direction although the evidence attribution is not manually verified.

\paragraph{Answer Set Sizes.} To ensure that relevance labels are correct and verifiable, we seek the help of crowdworkers. However, this meant that we needed to restrict the answer set sizes to 20 for the queries in our dataset, to make annotation feasible. On one hand, this is realistic for a search scenario because users may only be interested in a limited set of results. On the other hand, our dataset does not model a scenario where the answer set sizes are much larger.

\section*{Acknowledgements}
We would like to thank Isabel Kraus-Liang, Mahesh Maddinala, Andrew Smith, Daphne Domansi, and all the annotators for their work. We would also like to thank Mark Yatskar, Dan Roth, Zhuyun Dai, Jianmo Ni, William Cohen, Andrew McCallum, Shib Sankar Dasgupta and Nicholas Fitzgerald for useful discussions.

\bibliography{anthology,custom}
\bibliographystyle{acl_natbib}

\clearpage
\appendix

\section{Experimental Details and Hyperparameters}
\label{app:hyperparams}
All models were fine-tuned starting from T5 1.1 checkpoints~\footnote{\href{https://github.com/google-research/t5x/blob/main/docs/models.md}{https://github.com/google-research/t5x/blob/main/docs/models.md}}. We fine-tune T5 models on 32 Cloud TPU v3 cores\footnote{\url{https://cloud.google.com/tpu/}}. Fine-tuning takes less than 8 hours for all models.

\paragraph{Dual Encoder.} We used the t5x\_retrieval library~\footnote{\href{https://github.com/google-research/t5x_retrieval}{https://github.com/google-research/t5x\_retrieval}} for implementing dual encoder models. We tuned some parameters based on results on the validation set. Relevant hyperparameters for training the dual encoder are:
\begin{itemize}
    \item Learning Rate: 1e-3
    \item Warmup Steps: 1500
    \item Finetuning Steps: 15000
    \item Batch Size: 512
    \item Max Query Length: 64
    \item Max Candidate Length: 512
\end{itemize}

\paragraph{Classifier.} 

For negative examples, we sampled 250 random non-relevant documents and sampled 250 non-relevant documents from the top-1000 documents retrieved by BM25. We also replicated each positive example 50 times. We found an approximately even number of positive and negative examples lead to better performance than training with a large class imbalance. We found a combination of random negatives and negatives from BM25 performed better than using only either individual type of negative examples. Additionally, selecting negative examples from BM25 performed better than selecting negative examples from the T5-Large dual encoder.

For the T5 input we concatenated the query and truncated document text. The T5 output is the string ``relevant'' or ``not relevant''. To classify document relevance at inference time, we applied a threshold to the probability assigned to the ``relevant'' label, which we tuned on the validation set. When classifying BM25 candidates we used a threshold of $0.9$ and when classifying the dual encoder candidates we used a threshold of $0.95$.

Other relevant hyperparameters for training the classifier are:
\begin{itemize}
    \item Learning Rate: 1e-3
    \item Warmup Steps: 1000
    \item Finetuning Steps: 10000
    \item Batch Size: 1024
    \item Max Source Length: 512
    \item Max Target Length: 16
\end{itemize}

\section{Set Difference and Recall}
\label{app:sets_analysis}
\paragraph{Notation and Assumptions}
Let us assume we have two sets derived from the Wikipedia category graph, $\hat{A}$ and $\hat{B}$. The Wikipedia category graph can be missing some relevant entities, such that $\hat{A} \subset A$ and $\hat{B} \subset B$, where $A$ and $B$ are interpreted as the hypothetical sets containing \emph{all} relevant entities. We quantify the degree of missing entities by denoting recall as $r_A$ and $r_B$, such that $|\hat{A}| = r_A * |A|$ and $|\hat{B}| = r_B * |B|$. We quantify the fraction of elements in $A$ that are also in $B$ as $r_\cap$, such that $|A \cap B| = r_\cap * |A|$. For simplicity, we also assume that the overlap between $\hat{A}$ and $\hat{B}$ is such that $|\hat{A} \cap B| = r_A * |A \cap B|$ and $|\hat{A} \cap \hat{B}| = r_A * r_B * |A \cap B|$.

\paragraph{Derivation}
\emph{What is the recall ($r$) and precision ($p$) of $\hat{A} \setminus \hat{B}$ relative to $A \setminus B$ as a function of $r_A$, $r_B$, and $r_\cap$?}

First, we derive this function for recall:\footnote{We note some useful properties of pairs of sets $X$ and $Y$:
$X \setminus Y$ = $X \cap Y^c$, $|X \setminus Y| = |X| - |X \cap Y|$, if $X \subset Y$ then $X \cap Y = X$, and if $X \subset Y$ then $Y^c \subset X^c$.}

$$ r = \frac{|(A \setminus B) \cap (\hat{A} \setminus \hat{B})|} {|(A \setminus B)|} $$
$$ r = \frac{|(\hat{A} \setminus B)|} {|(A \setminus B)|} $$

$$ r = \frac{|\hat{A}| - |\hat{A} \cap B|}{|A| - |A \cap B|} $$
$$ r = \frac{r_A * |A| - r_A * r_\cap * |A|}{|A| - (r_\cap * |A|)} $$

$$ r = \frac{r_A * (1 - r_\cap) * |A|} {(1 - r_\cap) * |A|} $$
$$ \boxed{r = r_A} $$

And for precision:

$$ p =  \frac{|(A \setminus B) \cap (\hat{A} \setminus \hat{B})|} {|(\hat{A} \setminus \hat{B})|} $$
$$ p =  \frac{|(\hat{A} \setminus B)|} {|(\hat{A} \setminus \hat{B})|} $$

$$ p = \frac{|\hat{A}| - |\hat{A} \cap B|} {|\hat{A}| - |\hat{A} \cap \hat{B}|} $$

$$ p = \frac{r_A * |A| - r_A * r_\cap * |A|} {r_A * |A| - r_A * r_B * r_\cap * |A|} $$

$$ p = \frac{r_A * (1 - r_\cap) * |A|} {r_A * (1 - r_B * r_\cap) * |A|} $$
$$ \boxed{ p = \frac{(1 - r_\cap)} {(1 - r_B *  r_\cap)}} $$

\paragraph{Discussion}
While recall is simply equal to $r_A$, precision is a more complicated function of $r_B$ and $r_\cap$, and can be very low for large values of $r_\cap$. Intuitively, if subtracting $\hat{B}$ from $\hat{A}$ removes most of $\hat{A}$, then the precision of the resulting set will be dominated by the relevant entities missing from $\hat{B}$. This motivates limiting the intersection of the two sets used to construct queries involving set intersection. For example, if $r_B = 0.95$, then with $r_\cap < 0.8$, we can ensure $p > 0.83$.

\section{Annotation Details}
\label{app:annotation}

\begin{figure*}
    \centering
    \includegraphics[width=\textwidth]{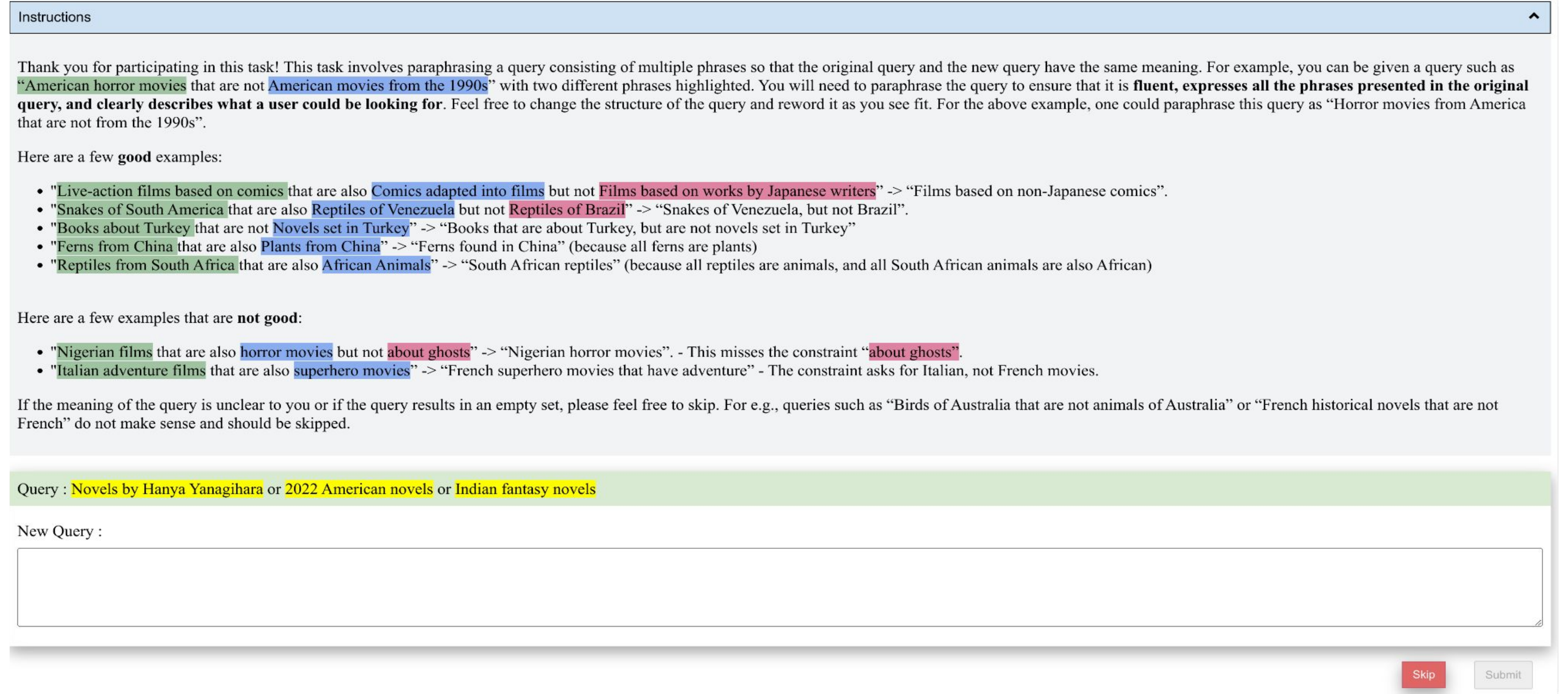}
    \caption{Annotation interface for the paraphrasing stage.}
    \label{fig:paraphrasing_interface}
\end{figure*}

\begin{figure*}
    \centering
    \includegraphics[width=\textwidth]{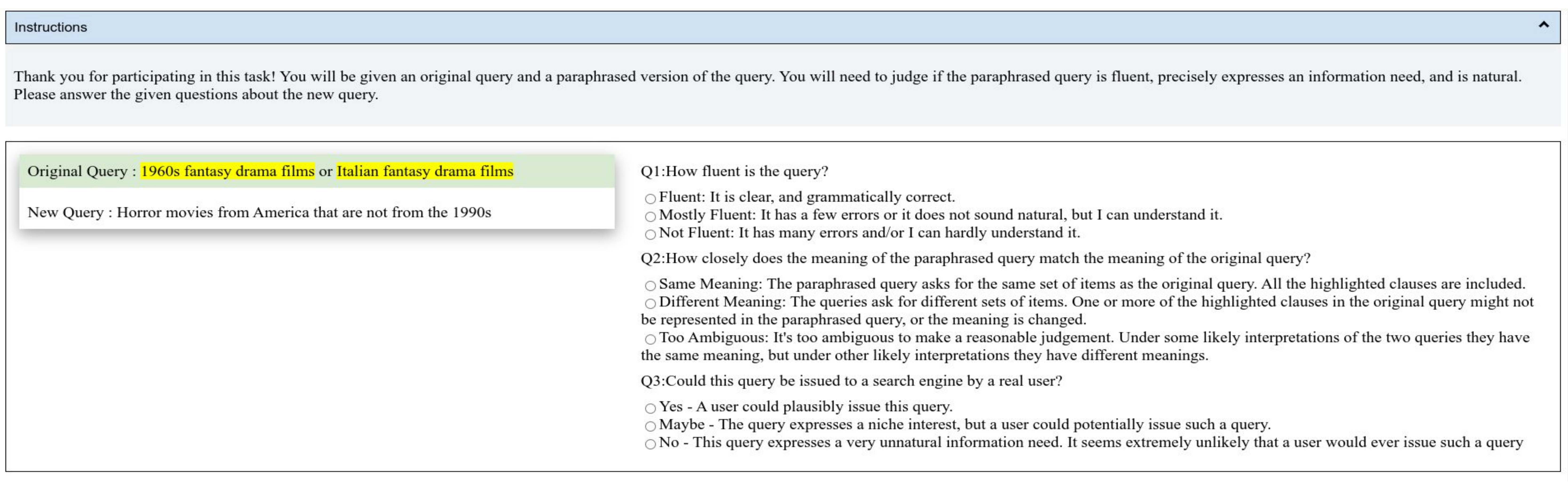}
    \caption{Annotation interface for the validation stage.}
    \label{fig:validation_interface}
\end{figure*}

\begin{figure*}
    \centering
    \includegraphics[width=\textwidth]{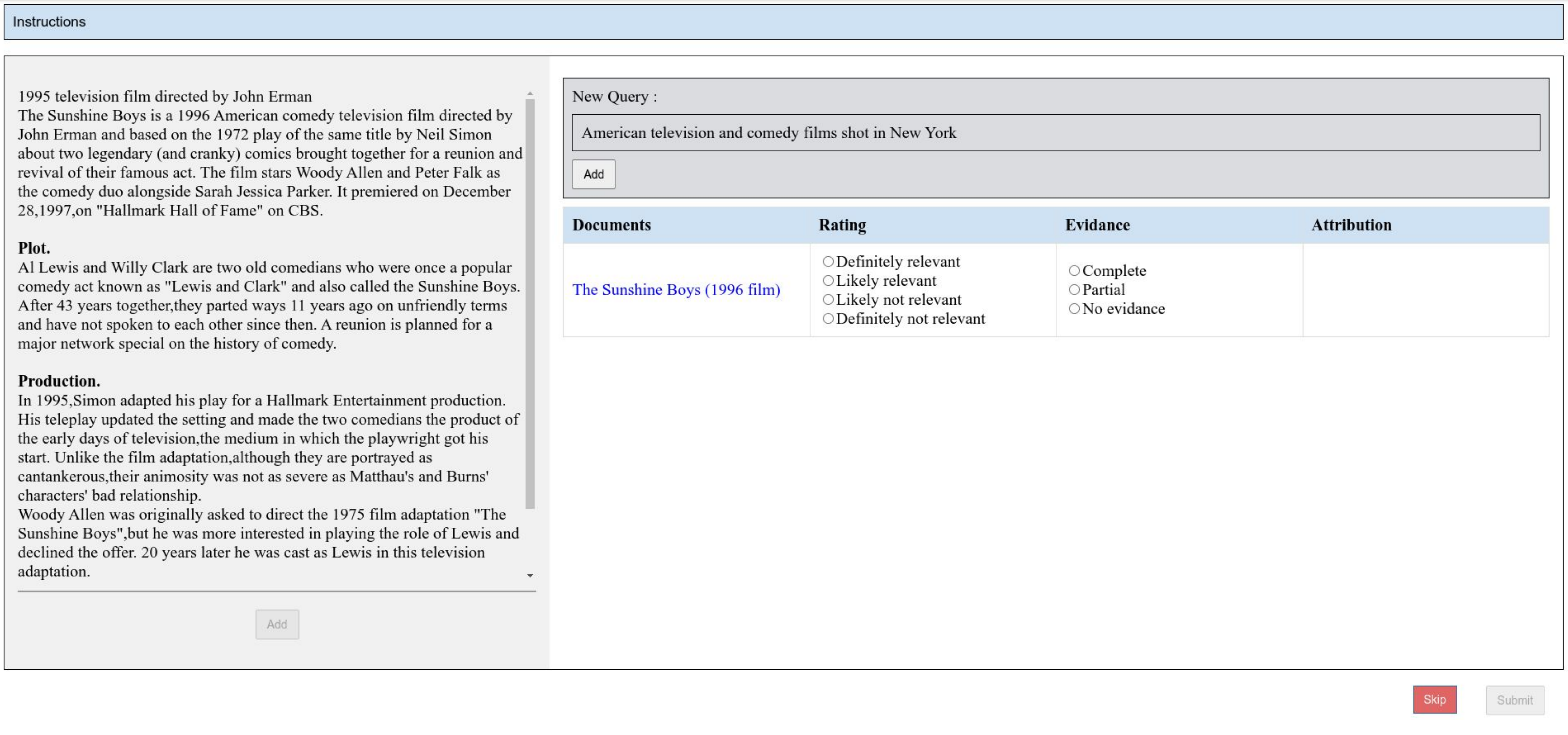}
    \caption{Annotation interface for the relevance labeling stage.}
    \label{fig:relevance_interface}
\end{figure*}

The annotation tasks in \textsc{Quest} were carried out by participants who were paid contractors. They are based in Austin, TX and either have a bachelor's degree (55\%) or equivalent work experience (45\%). They were paid by the hour for their work and were recruited from a vendor who screened them for knowledge of US English. They were informed of how their work would be used and could opt out. They received a standard contracted wage, which complies with living wage laws in their country of employment. The annotation interfaces presented to the annotators are shown in Figures \ref{fig:paraphrasing_interface}, \ref{fig:validation_interface} and \ref{fig:relevance_interface}.

\section{Impact of Memorization of Pre-training Data}
\label{app:memorization}

Since the T5 checkpoints we use to initialize our models were pre-trained on the C4 corpus (which includes Wikipedia), we investigate whether these models have memorized aspects of the Wikipedia category graph. We compare recall of the T5-based dual encoder model for Wikipedia documents that were created prior to the pre-training date of the T5 checkpoint compared with documents that were added after pre-training. We report these in Table~\ref{tab:pretraining_res}, along with the recalls for the same sets of documents with a BM25 retriever, for a baseline comparison. We note that the ratio of scores between the documents added before pre-training to documents added after pre-training is similar for both systems, which suggests factors other than memorization may explain the difference. For example, the documents created before vs. after the pre-training date have average lengths of 759.7 vs. 441.2 words, respectively.

\begin{table}[t!]
\centering
\scalebox{0.82}{
\begin{tabular}{lcc}
\toprule
& \multicolumn{2}{c}{Avg. Recall@100} \\
 \cmidrule(lr){2-3}
Retriever & Before & After \\
\midrule
BM25 & 0.183 & 0.050 \\
T5-Large DE & 0.466 & 0.171 \\
\bottomrule
\end{tabular}}

\caption{Average recall@100 on the subsets of documents created before vs after T5 pre-training.}
\label{tab:pretraining_res}
\end{table}

\end{document}